\documentclass{article}
\usepackage{spconf,amsmath,graphicx}

\usepackage{enumitem}
\setlist{nosep, leftmargin=14pt}
\usepackage{titlesec}
\usepackage{mwe} % to get dummy images
\usepackage{amsfonts}
\usepackage{booktabs}
\usepackage{multirow}
\title{A Calibrated Memorization Index (MI) for Detecting Training Data Leakage in Generative MRI Models}
\name{%
\parbox{\linewidth}{\centering
Yash Deo$^1$, Yan Jia$^1$, Toni Lassila$^2$, Victoria J Hodge$^1$,\\
{\it Alejandro F Frangi}$^{5,6}$, {\it Chenghao Qian}$^{2}$, {\it Siyuan Kang}$^4$, {\it Ibrahim Habli}$^1$
}%
}

\address{
        $^1$ Department of Computer Science, University of York, York, UK \\
        $^2$ School of Computer Science, University of Leeds, Leeds, UK \\
        $^4$ Department of Computing and Mathematics, Manchester Metropolitan University, Manchester, UK    \\   
        $^5$ Department of Computer Science, University of Manchester, Manchester, UK \\ 
        $^6$ Department of Cardiovascular Sciences, KU Leuven, Leuven, Belgium  
}

\begin{document}
%\ninept
\maketitle
\begin{center}
\footnotesize
© 2026 IEEE. Personal use of this material is permitted. Permission from IEEE must be obtained
for all other uses, in any current or future media, including reprinting/republishing this material for
advertising or promotional purposes, creating new collective works, for resale or redistribution to servers
or lists, or reuse of any copyrighted component of this work in other works.
\end{center}
\begin{abstract}
Image generative models are known to duplicate images from the training data as part of their outputs, which can lead to privacy concerns when used for medical image generation. We propose a calibrated per-sample metric for detecting  memorization and duplication of training data. Our metric uses image features extracted using an MRI foundation model, aggregates multi-layer whitened nearest-neighbor similarities, and maps them to a bounded \emph{Overfit/Novelty Index} (ONI) and \emph{Memorization Index} (MI) scores. Across three MRI datasets with controlled duplication percentages and typical image augmentations, our metric robustly detects duplication and provides more consistent metric values across datasets. At the sample level, our metric achieves near-perfect detection of duplicates. 
\end{abstract}

\section{Introduction}

Deep learning models for synthetic image generation are increasingly used to address data scarcity, class imbalance, and patient privacy concerns~\cite{khosravi2024synthetically,qian2024allweather}. However, these models can exhibit problematic behaviors, such as memorizing training data (data leakage)~\cite{akbar2025}. This risk is especially concerning in medical applications, as it can compromise patient privacy by reproducing personally identifiable data.

Validating the safety of these generative models requires the use of no-reference image-quality metrics (NRIQMs), since paired ground-truth images are typically unavailable for direct comparison. Distributional fidelity metrics such as FID and MMD~\cite{FID}, while useful for assessing visual quality, are often \emph{counter-diagnostic} for leakage. When duplicated training examples contaminate an evaluation set, the empirical test distribution moves closer to the training distribution, causing these scores to \emph{improve} even as memorization increases~\cite{MetricsMatter}.

Metrics designed specifically for duplication detection, such as the CT-score~\cite{Ct}, authenticity metric~\cite{Autptch}, or Vendi score~\cite{vendi}, also prove inadequate. These metrics are typically developed using features from models (like InceptionV3) trained on \emph{natural images}. This feature space transfers poorly to medical images, which are characterized by high anatomical regularity and limited appearance diversity. In medical images, clinically relevant variation is often subtle and localized. As a result, metrics based on natural-image features tend to misinterpret acquisition noise as ``diversity" or are easily fooled by simple augmentations, failing to expose near-duplicates~\cite{MetricsMatter}.

Furthermore, comparing different no-reference image quality metrics (NRIQMs) is complicated by arbitrary scaling effects that vary across datasets. An ideal metric for distribution-level comparisons should be robust to small perturbations (e.g., noise, small rotations) and provide a calibrated, consistent range of values to be interpretable (e.g., a score near 0 indicating novelty and a score near 1 indicating an exact copy) across different datasets. To our knowledge, no existing NRIQM for medical data memorization satisfies all these criteria.

We  propose a novel, calibrated, multi-scale metric for detecting data memorization in MRI datasets. Our method uses features extracted from a domain-specific MRI foundation model (MRI-CORE~\cite{MRI-CORE}) at multiple scales, capturing both fine-grained textures and gross anatomical structures. We validate our metric on three public MRI datasets (brain, knee, and spine)~\cite{Zhou2025CSpineSeg,brats1,ThighMRI} and show that it detects data leakage more consistently than generic metrics and is significantly more robust to common data augmentations. While tested on MRI, our methodology is general and applicable to other medical imaging modalities, provided an appropriate domain-specific feature space is used. The implementation code is available at :- https://github.com/YashDeo-York/Mem-index

\section{Methodology}

\begin{figure}[!ht]
\centerline{\includegraphics[width=0.5\textwidth]{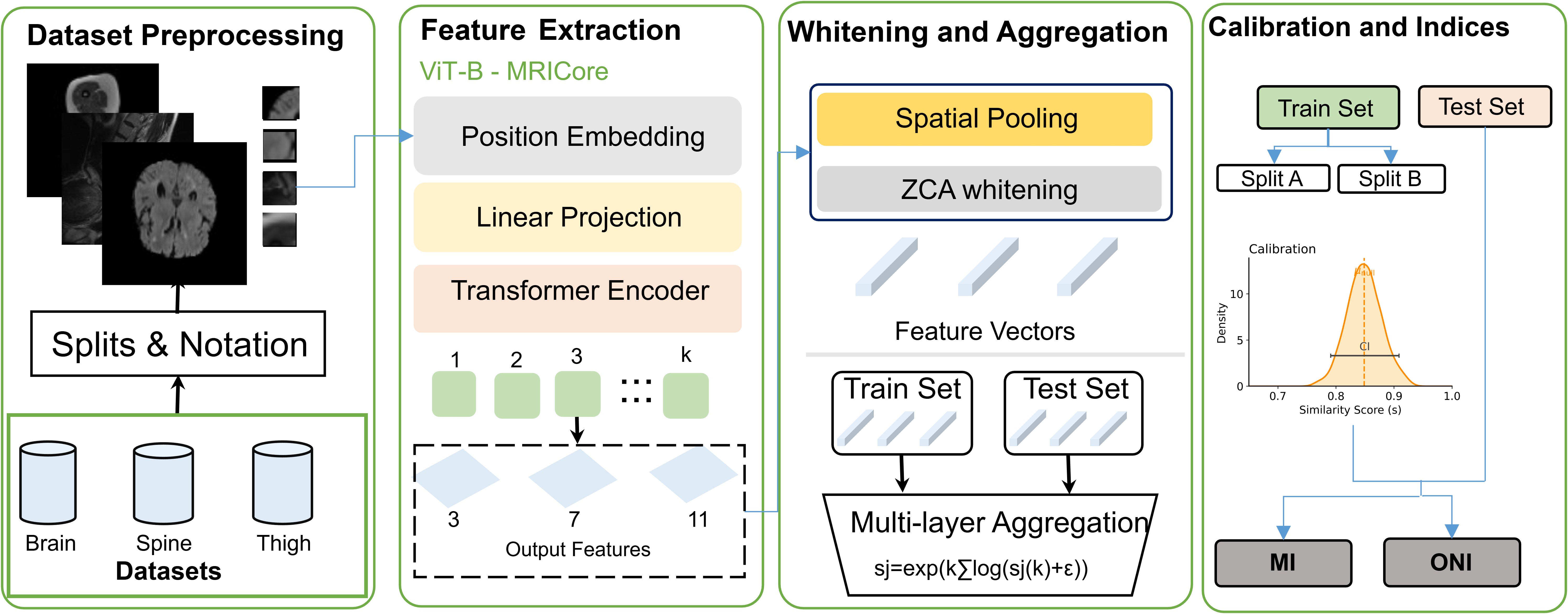}}
\caption{Overview of our methodology to calculate the Memorisation Index (MI) and the Overfit-Novelty Index (ONI)}
\label{fig1}
\end{figure}

We introduce a calibrated, multi-scale metric for slice-level memorisation detection in MRI datasets. The pipeline comprises: (i) multi-layer transformer feature extraction, (ii) per-layer whitening and similarity computation, (iii) multi-scale aggregation and (iv) calibration via empirical null distribution. A high level overview of the Methodology is shown in Figure ~\ref{fig1}.

\label{method}

\subsection{Feature Extraction}

We employ the MRI-CORE ~\cite{MRI-CORE} foundation model (SAM ViT-B encoder) pretrained on diverse MRI datasets. In ViT, early layers tend to exhibit localized attention similar to early convolutions in CNNs, while deeper layers progressively increase their 'attention distance' to integrate information more globally ~\cite{dosovitskiy2020image}; Hence , we extract features from transformer blocks $k \in \{3, 7, 11\}$, (from 11 total blocks) capturing representations from early texture patterns to high-level anatomical structures. For block $k$, spatial token outputs are reshaped to $\mathbf{F}^{(k)} \in \mathbb{R}^{C_k \times h_k \times w_k}$ and spatially pooled via adaptive average pooling:
\begin{equation}
    \mathbf{f}^{(k)} = \frac{1}{h_k w_k} \sum_{u,v} \mathbf{F}^{(k)}[:, u, v] \in \mathbb{R}^{C_k}
\end{equation}

We denote train/test pooled features by $\mathcal{F}^{(k)}_{\text{train}} = \{\mathbf{f}^{(k)}_i\}$ and $\mathcal{F}^{(k)}_{\text{test}} = \{\mathbf{g}^{(k)}_j\}$.

\subsection{Whitening and Layer-wise Similarity}

To render cross-layer similarities comparable, we apply ZCA whitening ~\cite{Kessy_2018} estimated on train features per layer. ZCA whitening decorrelates features while maintaining the original feature space, unlike PCA which rotates to principal components. With mean $\boldsymbol{\mu}^{(k)}$ and covariance $\mathbf{C}^{(k)}$:
\begin{equation}
    \hat{\mathbf{f}}^{(k)} = (\mathbf{f}^{(k)} - \boldsymbol{\mu}^{(k)})\mathbf{W}^{(k)}, \quad \mathbf{W}^{(k)} = (\mathbf{C}^{(k)} + \epsilon \mathbf{I})^{-1/2}
\end{equation}
where $\epsilon = 10^{-6}$ ensures numerical stability. Whitened vectors are $\ell_2$-normalized. For each test sample $j$ at layer $k$, we compute the maximum cosine similarity to any training sample:
\begin{equation}
    s^{(k)}_j = \max_i \left\langle \frac{\hat{\mathbf{g}}^{(k)}_j}{\|\hat{\mathbf{g}}^{(k)}_j\|_2}, \frac{\hat{\mathbf{f}}^{(k)}_i}{\|\hat{\mathbf{f}}^{(k)}_i\|_2} \right\rangle
\end{equation}

\subsection{Multi-Scale Aggregation}
\label{sec:multiscale}

Given the per-layer nearest–neighbour similarities $s^{(k)}_j\!\in[0,1]$ computed above for layers $k\!\in\!K$ (here $K=\{3,7,11\}$), we aggregate them into a single sample-level score via a geometric mean:
\begin{equation}
  s_j 
  \;=\;
  \Bigg(\prod_{k\in K}\big(s^{(k)}_j+\epsilon\big)\Bigg)^{\!1/|K|}
  \;=\;
  \exp\!\Bigg(\frac{1}{|K|}\sum_{k\in K}\log\big(s^{(k)}_j+\epsilon\big)\Bigg),
\end{equation}
and define the distance $d_j = 1 - s_j$ and set $\epsilon=10^{-6}$.

This log–average acts as a product-of-experts, rewarding \emph{cross-scale consensus}: a sample is scored as highly similar only if it is simultaneously close across early, mid, and late features. 
Relative to an arithmetic mean, it suppresses single-layer outliers and penalizes cases where any scale fails to find a close neighbor. 
For diagnostics, we also retain per-layer neighbor indices and a consensus count (number of layers selecting the same neighbor).
\subsection{Calibration and Memorization Index}

To establish significance thresholds, we construct an empirical null distribution by bootstrapping the training set. We randomly sample independent subsets $A, B$ (50\% of training data, $n=10$ iterations) and compute aggregated similarities $s_{A,B}$ using the same whitening and aggregation pipeline. The null parameters are:
\begin{equation}
    \mu_{\text{null}} = \mathbb{E}[s_{A,B}], \quad \sigma_{\text{null}} = \sqrt{\text{Var}(s_{A,B}) + 10^{-8}}
\end{equation}

Test similarities are standardized into a \textit{Memorization Index} (MI) and mapped to a bounded \textit{Overfit/Novelty Index} (ONI):
\begin{equation}
    \text{MI}_j = \frac{s_j - \mu_{\text{null}}}{\sigma_{\text{null}}}, \quad \text{ONI}_j = -\tanh(\text{MI}_j)
\end{equation}

Values of ONI close to $-1$ indicate high similarity (potential memorisation), values of ONI close to $0$ indicate similarity consistent with the null distribution (i.e., the typical, expected similarity between two unrelated samples from the training set) and values close to $+1$ indicate novelty.

\section{Results}
\label{sec:results}

\noindent {\bf Experimental Setup:}
We evaluated our metric on three MRI datasets: BRATS~\cite{brats2} brain tumour images (axial), knee/thigh images~\cite{ThighMRI} (sagittal), and spine images (sagittal) ~\cite{Zhou2025CSpineSeg}, each with 500 randomly selected slices divided into disjoint train and test sets. We introduced controlled duplication by replacing test samples with randomly selected training slices at 5\%, 15\%, 30\%, and 45\% duplication levels. To assess robustness, the duplicates were perturbed with eight augmentations: Gaussian noise ($\sigma \in \{0.01, 0.02\}$), rotations ($\pm3°, \pm5°$), horizontal/vertical flips, and intensity scaling ([0.9, 1.1]). We tested the ability of our metrics (MI and ONI) to detect this duplication and compared them against CT Score~\cite{Ct}, FID, MMD, AuthPct~\cite{Autptch}, and Vendi Score~\cite{vendi}. 

\begin{figure*}[t]
  \centering
  \includegraphics[width=0.91\textwidth,height=0.5\linewidth]{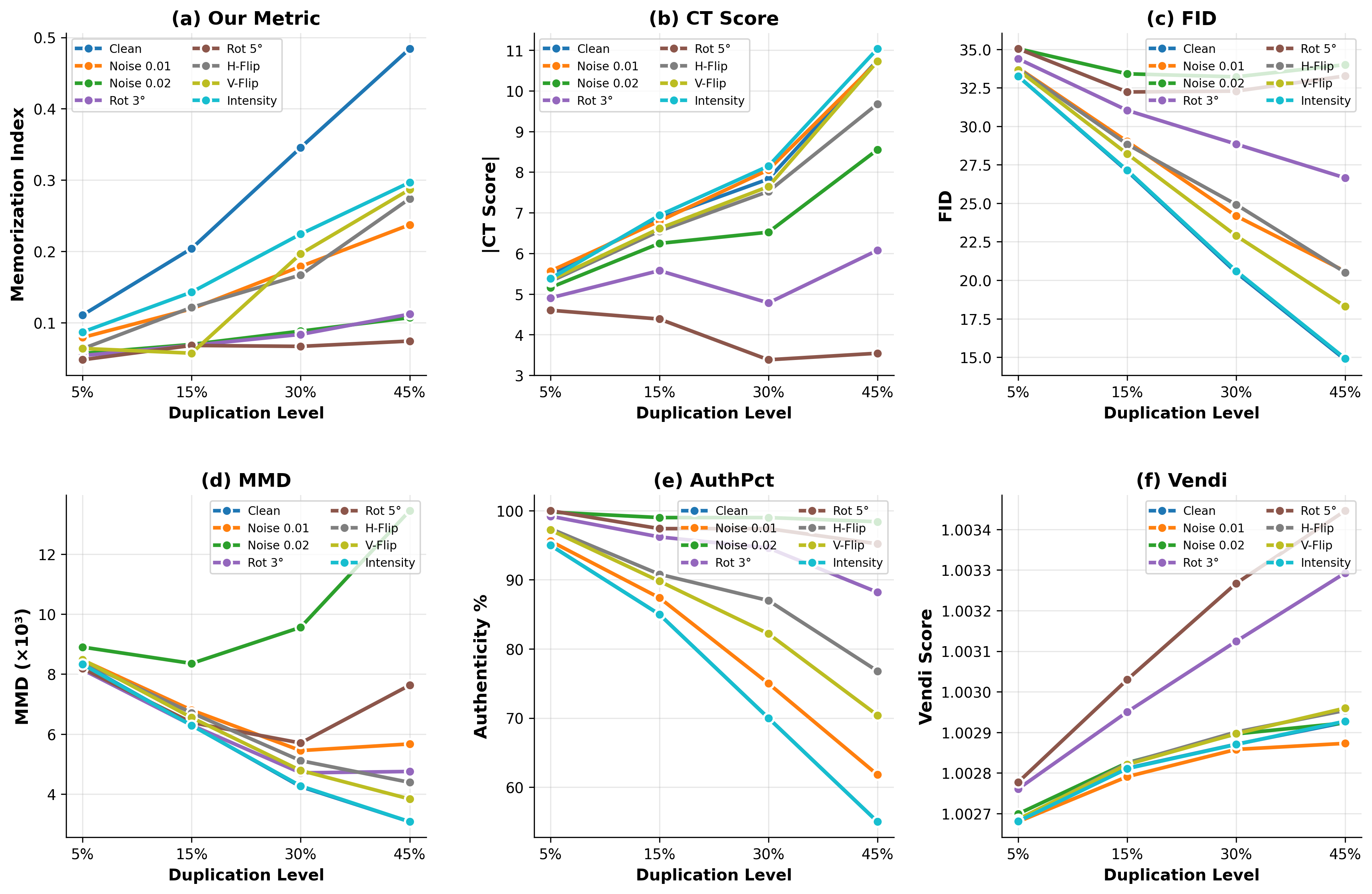}
  \caption{\textbf{Metric response to duplication under augmentations.}
  (a) MI increases near-linearly and remains tight across augmentations.
  (b) CT spreads across augmentations, especially at higher duplication percentages.
  (c--d) FID/MMD decrease as duplication increases, which can be misleading if used as the only quality metric.
  (e) AuthPct is highly augmentation-sensitive.
  (f) Vendi shows little/no signal.
 }
  \label{fig:metric_trends}
  \vspace{-2mm}
\end{figure*}

\noindent {\bf Metric responses to duplication under augmentations:}
Fig.~\ref{fig:metric_trends} shows the overall trends of each metric as the duplication level is increased for the different types of augmentations applied. In general, every metric responded in some way to increased duplication, with most metric exhibiting lower sensitivity as more noise or small rotations were applied. 
%We report mean and standard deviation of MI and ONI stratified by duplication status (clean vs. memorized). 

 %Cross-layer consensus measures agreement on nearest neighbors across scales, indicating robustness of detection. I THINK THIS WAS REMOVED AT SOME POINT
 
 %\noindent\textbf{Setup.}
%We evaluate five set-level metrics---\textbf{FID}, \textbf{MMD}, \textbf{CT score}, \textbf{Vendi}, and \textbf{AuthPct}---and our method (``\textbf{Ours}'' distance and its calibrated per-sample \textbf{ONI}) on three MRI datasets (BRATS, Knee, Spine) under controlled leakage levels (5/15/30/45\%) and eight augmentations.

\noindent\textbf{Interpretation of observed trends:}
As the percentage of duplication increases, MI increases nearly linearly even in the presence of augmentations (Fig.~\ref{fig:metric_trends}a). CT score also increases, but is not always monotonic and exhibits a stronger spread when augmentations are applied (Fig.~\ref{fig:metric_trends}b). FID/MMD decrease with higher duplication (Fig.~\ref{fig:metric_trends}c--d), which could falsely imply better image fidelity. The Vendi score provides little/no usable signal (Fig.~\ref{fig:metric_trends}f), while AuthPct detects duplication well but is highly sensitive to the augmentation applied (Fig.~\ref{fig:metric_trends}e).

%; therefore fidelity metrics \emph{must be paired} with a duplication detector to avoid false conclusions about generation quality. THIS BELONGS IN THE CONCLUSIONS

\noindent\textbf{Robustness over different augmentations.} To measure how consistent the metrics are to different image augmentations, we computed statistics for the same score over different augmentations. Table~\ref{tab:robustness} quantifies the augmentation spread (stdev across the 8 augmentations at each duplication level). MI is 6--20$\times$ more stable than CT on all three datasets (e.g., at 45\% leak on BRATS: 2.73 vs.\ 0.14; Spine: 0.85 vs.\ 0.14; Knee: 0.92 vs.\ 0.14), making it easier to interpret consistently.

\begin{table}[t]
\centering
\caption{Robustness to augmentations: standard deviation across 8 augmentation types at each duplication level. Our metric is 6--20$\times$ more stable than CT Score.}
\label{tab:robustness}
\small
\begin{tabular}{c|cc|cc|cc}
\toprule
\multirow{2}{*}{\textbf{Dup}} & \multicolumn{2}{c|}{\textbf{BRATS}} & \multicolumn{2}{c|}{\textbf{Spine}} & \multicolumn{2}{c}{\textbf{Knee}} \\
& \textbf{CT} & \textbf{Ours} & \textbf{CT} & \textbf{Ours} & \textbf{CT} & \textbf{Ours} \\
\midrule
5\%  & 0.32 & \textbf{0.02} & 0.18 & \textbf{0.02} & 0.14 & \textbf{0.02} \\
15\% & 0.87 & \textbf{0.05} & 0.47 & \textbf{0.05} & 0.39 & \textbf{0.05} \\
30\% & 1.75 & \textbf{0.09} & 0.70 & \textbf{0.10} & 0.69 & \textbf{0.09} \\
45\% & 2.73 & \textbf{0.14} & 0.85 & \textbf{0.14} & 0.92 & \textbf{0.14} \\
\midrule
\textbf{Ratio} & \multicolumn{2}{c|}{\textbf{18×}} & \multicolumn{2}{c|}{\textbf{9×}} & \multicolumn{2}{c}{\textbf{7×}} \\
\bottomrule
\end{tabular}
\vspace{-2mm}
\end{table}

\noindent\textbf{Cross-dataset consistency:} A metric whose scale shifts from dataset to dataset is hard to interpret. We performed the duplicate detection experiments in three different MRI datasets to test how consistent the metric values were. Table~\ref{tab:consistency} reports the coefficient of variation (CV) across datasets at each duplication level. MI achieves 5.5$\times$ lower CV  on average (0.072 vs.\ 0.395), with the largest gap at high levels of duplication (12$\times$ at 45\%). This potentially enables universal thresholds for duplication to be set, e.g., a CT value of ``$-0.3$'' is dataset-dependent, whereas MI or ONI has a stable range of interpretation. Table~\ref{tab:oni_calibration} showcases clean ONI having stability across multiple datasets. This consistency / stability of MI/ONI across datasets justifies their usage as early-leak detectors across datasets.

\begin{table}[t]
\centering
\caption{Cross-dataset consistency measured by coefficient of variation (CV). Our metric achieves 5.5$\times$ better consistency, enabling universal thresholds.}
\label{tab:consistency}
\small
\begin{tabular}{c|cc|c}
\toprule
\textbf{Duplication} & \textbf{CT Score} & \textbf{Our Metric} & \textbf{Improvement} \\
\midrule
5\%  & 0.683 & \textbf{0.163} & 4.2$\times$ \\
15\% & 0.441 & \textbf{0.069} & 6.4$\times$ \\
30\% & 0.225 & \textbf{0.035} & 6.4$\times$ \\
45\% & 0.229 & \textbf{0.019} & 12.0$\times$ \\
\midrule
\textbf{Mean} & 0.395 & \textbf{0.072} & \textbf{5.5$\times$} \\
\bottomrule
\end{tabular}
\vspace{-2mm}
\end{table}

\begin{table}[t]
\centering
\caption{ONI scores for clean samples: stable across all conditions (CV $<$ 0.013), enabling universal threshold (ONI $<$ 0.68).}
\label{tab:oni_calibration}
\small
\begin{tabular}{l|ccc}
\toprule
\textbf{Dataset} & \textbf{Mean} & \textbf{Std} & \textbf{CV} \\
\midrule
BRATS & 0.700 & 0.004 & 0.006 \\
Knee  & 0.710 & 0.009 & 0.013 \\
Spine & 0.735 & 0.007 & 0.010 \\
\midrule
\textbf{All} & \textbf{0.715} & \textbf{0.017} & \textbf{0.023} \\
\bottomrule
\end{tabular}
\vspace{-2mm}
\end{table}

\noindent\textbf{Detection of duplicate samples (MI only):}
Unlike most other metrics, our metric provides \emph{per-sample} scores and can be used identify (near) duplicate images. Detection performance was quantified via ROC-AUC and average precision (AP) using known duplication labels from synthetic contamination. Table~\ref{tab:auc_performance} shows AUC by augmentation: detection is perfect ($\approx$1.0) for clean, noise, and intensity; it remains good under geometric transforms (e.g., rot$\pm3^\circ$: mean 0.871; rot$\pm5^\circ$: 0.758; flips $\approx$0.73). This directly supports curation (flagging/removing a small set of memorized slices), which set-level metrics cannot address.

% TABLE 4: AUC performance (single column, compact)
\begin{table}[t]
\centering
\caption{Sample-level detection (AUC) by augmentation type. Perfect detection (1.0) for clean/noise/intensity; good detection ($>$0.72) for geometric transforms. %\textit{FID/MMD/CT cannot do this.}
}
\label{tab:auc_performance}
\small
\begin{tabular}{l|cc}
\toprule
\textbf{Augmentation} & \textbf{Mean AUC} & \textbf{Min AUC} \\
\midrule
Clean              & 1.000 & 1.000 \\
Noise (0.01)       & 1.000 & 1.000 \\
Noise (0.02)       & 1.000 & 0.999 \\
Intensity Scale    & 1.000 & 1.000 \\
Rotation (±3°)     & 0.871 & 0.805 \\
Rotation (±5°)     & 0.758 & 0.711 \\
H-Flip             & 0.733 & 0.626 \\
V-Flip             & 0.727 & 0.635 \\
\midrule
\textbf{Overall}   & \textbf{0.886} & \textbf{0.626} \\
\bottomrule
\end{tabular}
\vspace{-2mm}
\end{table}

\section{Discussion}

We introduced a metric for detecting memorization that: (i) is monotonic and augmentation-stable at the set-level, (ii) avoids the misleading directionality of fidelity metrics under duplication, (iii) maintains cross-dataset scale consistency that permits universal thresholds, and (iv)  provides per-sample scores to identify and remove leaked training images. Together, these properties turn memorization assessment from a passive diagnostic into a practical curation tool that complements fidelity metrics and improves the reliability of image generative models. Our experiments establish three practical properties of the proposed memorization detector. First, at the \emph{set level}, our distance increases monotonically with duplication and remains tight across common augmentations, matching the trend-signal of CT while being substantially more stable (Table~\ref{tab:robustness}) and more consistent across datasets (Table~\ref{tab:consistency}). 
This matters because a metric whose scale drifts by dataset is hard to interpret; CT’s starting points vary widely, whereas our score (MI and ONI) exhibit small cross-dataset CV, enabling universal thresholds.

Second, \emph{fidelity} metrics (FID/MMD) decrease as duplication increases, which by construction suggests ``better quality'' even when the generator is copying.
This confirms that fidelity should be paired with a memorization metric. 
Vendi shows negligible sensitivity and AuthPct is highly augmentation-dependent, limiting their diagnostic value.

Third, unlike set-level metrics, our method yields \emph{per-sample} scores and can recover the actual near-duplicate pairs.
Across datasets, ONI achieves near-perfect AUC for clean/noise/intensity cases and remains effective under small geometric changes (Table~\ref{tab:auc_performance}), making data curation \emph{actionable} (flag/remove a tiny subset) rather than merely observational.

\paragraph*{Limitations and failure modes.}
Sensitivity under small rotations and flips is lower than for appearance-preserving perturbations, especially at very low leak (5\%). 
This is unsurprising given the local-invariance profile of the embeddings and the calibration’s null. 
Two lightweight remedies are straightforward: (i) rotation/translation pooling of features before nearest-neighbor aggregation, and (ii) incorporating mild geometric jitter into the empirical null for MI$\!\to\!$ONI calibration.
% Both retain the clean advantages (set-level monotonicity and cross-dataset stability) while improving geometric robustness.
% A second limitation is computational: nearest-neighbour search scales with the reference pool; however, approximate indexing (e.g., FAISS IVF-PQ/HNSW) makes million-scale evaluation practical and does not change the conclusions.

% \paragraph*{Broader impact and scope.}
% Our study focuses on MRI, but the ingredients—multi-layer features, whitening, geometric aggregation, and calibrated ONI—are modality-agnostic.
% The cross-dataset stability we observe (Table~\ref{tab:consistency} and Table~\ref{tab:oni_calibration}) suggests the approach may transfer with minimal threshold tuning, which we leave to future work on natural images and text-to-image synthesis.

%\section{Conclusion}

\paragraph{Acknowledgements:} This work was supported by the Centre for Assuring Autonomy, a partnership between Lloyd’s Register Foundation and the University of York. AFF acknowledges support from the Royal Academy of Engineering under the RAEng Chair in Emerging Technologies (INSILEX CiET1919/19), ERC Advanced Grant– UKRI Frontier Research Guarantee (INSILICO EP/Y030494/1), the UK Centre of Excellence on in-silico Regulatory Science and Innovation (UK CEiRSI) (10139527), the National Institute for Health and Care Research (NIHR) Manchester Biomedical Research Centre (BRC) (NIHR203308), the BHF Manchester Centre of Research Excellence (RE/24/130017), and the CRUK RadNet Manchester (C1994/A28701).

\paragraph{Compliance with Ethical Standards:} The experiments in this study were based on publicly available data collected with informed written consent and anonymised before distribution. The authors report no financial conflicts of interest.

\bibliographystyle{IEEEbib}
\bibliography{strings,refs}

\end{document}